\documentclass[letterpaper, 10 pt, conference]{ieeeconf}  

\IEEEoverridecommandlockouts                              

\usepackage{graphics} 
\usepackage{epsfig} 
\usepackage{mathptmx} 
\usepackage{times} 
\usepackage[font={small,it}]{caption}
\usepackage{hyperref}
\usepackage{amsmath}
\usepackage{amssymb}  

\DeclareMathAlphabet{\mathcal}{OMS}{cmsy}{m}{n}

\newcommand{\robot}{\mathcal{R}}
\newcommand{\task}{\mathcal{T}}

\newcommand{\universe}{\mathcal{U}}
\newcommand{\test}{\text{test}}

\long\def\ignorethis#1{}

\renewcommand{\eqref}[1]{Equation~(\ref{#1})}

\DeclareGraphicsExtensions{.pdf,.png,.jpg}

\usepackage{algorithm}
\usepackage{algpseudocode}
\algtext*{EndFor}
\usepackage{subfig}

\title{\LARGE \bf
Learning Modular Neural Network Policies for Multi-Task and Multi-Robot Transfer
}

\author{Coline Devin$^*$$^{1}$\thanks{$^*$The first two authors contributed equally to this work.} \and Abhishek Gupta$^*$$^{1}$\and Trevor Darrell$^{1}$ \and Pieter Abbeel$^{1}$  \and Sergey Levine$^{1}$
\thanks{$^{1}$Berkeley Artificial Intelligence Research, Department of Electrical Engineering and Computer Science, University of California, Berkeley}%
}

\begin{document}

\maketitle
\thispagestyle{empty}
\pagestyle{empty}

\begin{abstract}
Reinforcement learning (RL) can automate a wide variety of robotic skills, but learning each new skill requires considerable real-world data collection and manual representation engineering to design policy classes or features. Using deep reinforcement learning to train general purpose neural network policies alleviates some of the burden of manual representation engineering by using expressive policy classes, but exacerbates the challenge of data collection, since such methods tend to be less efficient than RL with low-dimensional, hand-designed representations.
Transfer learning can mitigate this problem by enabling us to transfer information from one skill to another and even from one robot to another.
We show that neural network policies can be decomposed into ``task-specific'' and ``robot-specific'' modules, where the task-specific modules are shared across robots, and the robot-specific modules are shared across all tasks on that robot.
This allows for sharing task information, such as perception, between robots and sharing robot information, such as dynamics and kinematics, between tasks. We exploit this decomposition to train mix-and-match modules that can solve new robot-task combinations that were not seen during training. Using a novel neural network architecture, we demonstrate the effectiveness of our transfer method for enabling zero-shot generalization with a variety of robots and tasks in simulation for both visual and non-visual tasks. 
\end{abstract}


\section{Introduction}


Deep reinforcement learning (RL) has been successful in multiple domains, including learning to play Atari games~\cite{mnih}, simulated qnd real locomotion~\cite{trpo, igordarwin} and robotic manipulation~\cite{levinefinn16JMLR}. The onerous data requirements for deep RL make transfer learning appealing, but the policies learned by these algorithms lack clear structure, making it difficult to leverage what was learned previously for a new task or a new robot. The relationship between the optimal policies for different combinations of tasks and robots is not immediately clear, and doing transfer via finetuning does not work well for robotic RL domains due to the lack of direct supervision in the target domain. 

However, much of what needs to be learned for robotic skills (dynamics, perception, task steps) is decoupled between the task and the robot. Part of the information gained during learning would help a new robot learn the task, and part of it would be useful in performing a new task with the same robot. Instead of throwing away experience on past tasks, we propose learning structured policies that decompose in a way that we can use transfer learning to help a robot benefit from its own past experience on other tasks, as well as from the experience of other, morphologically different robots, to learn new tasks more quickly.


In this paper, we address the problem of transferring experience across different robots and tasks.
Specifically, we consider the problem of transferring information
across robots with varying morphology, including varying numbers of links and joints and across a diverse range of tasks. The discrepancy in the morphologies of the robots and the goals of the tasks prevents us from directly reusing policies learned on multiple tasks or robots for a new combination, and requires us to instead devise a novel modular approach to policy learning. An additional difficulty is determining which information can be transferred to a new robot and which can be transferred to a new task.
As an example, consider a robot that has learned how to fold shirts and pants. Given a new robot which when only knows how to fold shirts, we wish to transfer something about folding pants from the first to the second robot which combined with the second robot's experience folding shirts, would help it achieve the task. The first robot would transfer task information to the second, while the second robot would transfer its understanding of its own dynamics and kinematics from folding shirts to folding pants.

In this work, we explore modular decomposable policies that are amenable to cross-robot and cross-task transfer. By separating the learned policies into a task-specific and robot-specific component, we can train the same task-specific component across all robots, and the same robot-specific component across all tasks. The robot and task-specific modules can then be mixed and matched to execute new task and robot combinations or, in the case of particularly difficult combinations, jump start the learning process from a good initialization.
In order to produce this decomposition of policies into task-relevant and robot-relevant information, we show that policies represented by neural networks can be decomposed into task-specific or robot-specific modules. We demonstrate that these modules can be trained on a set of robot-task combinations and can be composed to enable zero-shot performance or significantly sped up learning for unseen robot-task combinations. 

Our contributions are as follows:
\begin{enumerate}
    \item We address robotic multi-task and transfer learning by training policy modules that are decomposed over robots and tasks, so as to handle novel robot-task combinations with minimal additional training.
    \item We analyze regularization techniques that force the modules to acquire a generalizable bottleneck interface.
    \item We present a detailed evaluation of our method on a range of simulated tasks for both visual and non-visual policies.
\end{enumerate}
To the best of our knowledge, this is the first method to decompose policy neural networks into interchangeable modules than can perform zero-shot transfer with novel module combinations.

\section{Related Work}
Robotic skill learning via reinforcement learning has been studied extensively in recent years \cite{survey_deisenroth,survey_kober,reps_peters,Schaal}, and transfer learning in particular has been recognized for some time as an important direction in robotic learning \cite{Taylor_stone_survey,Caruana95NIPS,Carauna97ML,Crummon02JAIR,Konidaris06ICML}, due to its potential for reducing the burden of expensive on-policy data collection for learning large repertoires of complex skills. \cite{Ramon07ECML} and \cite{Madden04AIR} transfer between tasks by storing symbolic knowledge in knowledge bases.
Work by Guestrin et al. learned to play many versions of a computer strategy game by decomposing the value function into different domains~\cite{GuestrinIJCAI03}. The PG-Ella algorithm uses policy gradients for sequential multitask learning~\cite{pgella}.
Past work in transfer on robotics domains includes shaping the target reward function from the source policy~\cite{konidaris07ijcai,Mihalkova09IJCAI} and learning a  mapping between tasks~\cite{taylor07jmlr}. Another transfer approach used by~\cite{Drummond02Jair} is to split each task into sub-tasks and transfer the sub-tasks between tasks. An early work by Caruana uses backpropagation to learn many tasks jointly~\cite{Caruana95NIPS}.
Our work differs from these prior methods in that we explicitly consider transfer across tasks with two factors of variation, which in our experiments are robot identity and task identity. This allows us to decompose the policy into robot-specific and task-specific modules, which perform zero-shot transfer by recombining novel pairs of modules. Our method is complementary to prior transfer learning techniques in that we address primarily the question of policy representation, while prior methods focus on algorithmic questions. 

Beyond robotic learning, recent work in computer vision and other passive perception domains has explored both transfer learning and recombination of neural network modules. Pretraining is a common transfer learning technique in deep learning~\cite{Decaf}. However, pretraining cannot provided zero-shot generalization, and finetuning is ill-defined outside of supervised learning. Domain adaptation techniques have been used to adapt training data in the face of systematic domain shift \cite{domain_adaptation}, and more recently, work on modular networks for visual question answering has been demonstrated with good results \cite{andreas_modular_nets}. Our method differs from these prior approaches by directly considering robotic policy learning, where the policy must consider both the invariances and task-relevant differences across domains.

Although our method is largely agnostic to the choice of policy learning algorithm, we use the guided policy search method in our experiments~\cite{levinefinn16JMLR}. This algorithm allows us to train high-dimensional neural network policy representations, which can be readily decomposed into multiple interconnected modules. Other recent work on high-dimensional neural network policy search has studied continuous control tasks for simulated robots \cite{trpo,ddpg}, playing Atari games \cite{mnih}, and other tasks \cite{minecraft}. Recent work on progressive neural network also proposes a representation suitable for transfer across Atari games \cite{Rusu16ArXiv}, but does not provide for zero-shot generalization to new domains, and work by Braylen et al. used evolutionary algorithms to recombine networks trained for different Atari games, but again did not demonstrate direct zero-shot generalization~\cite{DBLP:Braylan15ArXiv}. We further emphasize that our approach is not in fact specific to neural networks, and our presentation of the method describes a generic framework of composable policy modules that can easily be extended to other representations.

\begin{figure}[t!]
  \centering
  \includegraphics[height = 2cm, width=.75\linewidth]{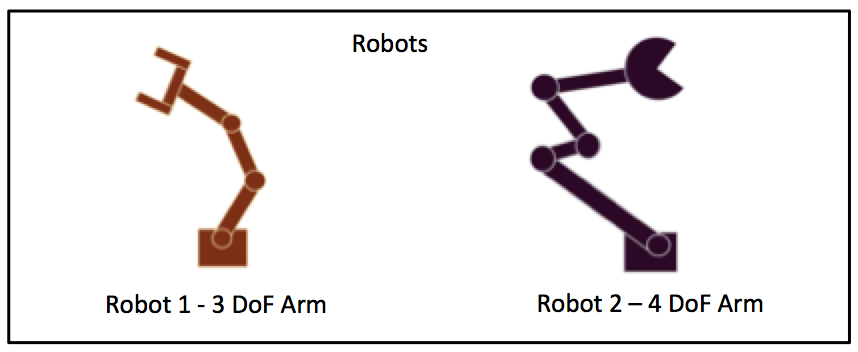}
  \includegraphics[height = 2cm, width=.75\linewidth]{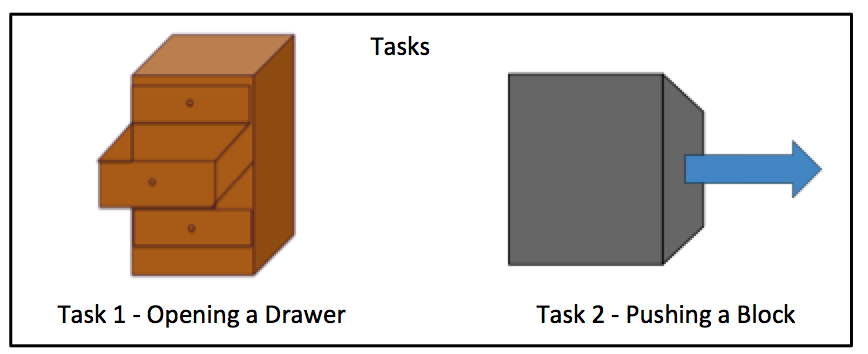}
  \caption{The 3DoF and a 4DoF robot which specify one degree of variation (robots) 
  in the universe described in Section~\ref{sec:modpolnet} as well as the tasks of opening a drawer and pushing a block which specify the other degree of variation (tasks) 
  in the universe.}
  \label{fig:robots}
\end{figure}


\begin{figure}[t!]
  \centering
  \includegraphics[height = 5cm, width=.75\linewidth]{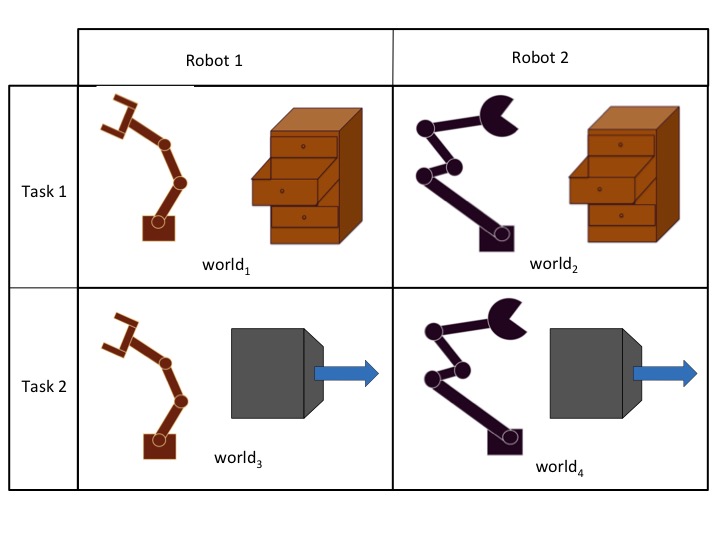}
  \caption{The possible worlds enumerated for all combinations of tasks and robots for the universe described in Section~\ref{sec:modpolnet}}
  \label{fig:possibleworlds}
\end{figure}
\section{Modular Policy Networks}
\label{sec:modpolnet}
The problem setting that this work addresses is enabling transfer across situations that can vary along some predefined discrete degrees of variation (DoVs). These DoVs can be different robot morphologies, different task goals, different object characteristics, and so forth. We define a ``world'' $w$
to be an instantiation of these DoVs, and our ``universe'' $\universe$ to be the set of all possible worlds. To illustrate this formalism, consider a universe with the following 2 DoVs: robot structure (3 DoF and 4 DoF), and task (open drawer and pushing a block). This universe would have 4 possible worlds: 3 DoF arm opening a drawer, 3 DoF arm pushing a block, 4 DoF arm opening a drawer, and 4 DoF arm pushing a block. 

Learning a single policy to act in all worlds is non-optimal because differences in the degrees of variation result in different optimal policies. Strategies required to push a block are quite different from those for a drawer, and robots with different numbers of joints would require different controls. 

Standard reinforcement learning algorithms would treat each world $w$ as a separate problem and learn an optimal policy for that world from scratch. However, the worlds have overlap: although a 3 DoF arm pushing a block and a 3 DoF arm opening a drawer are doing different tasks, they share the same arm, and thus will have commonalities in their optimal policy. This concept can be extended to other degrees of variation, such as when different arms perform the same task, or when the same arm interacts with different objects. Using this notion of commonality between worlds that share some DoVs, we tackle the problem of training policies for a subset of all worlds in a universe, and use these to enable fast transfer to new unseen worlds. Our experiments operate in the regime of 2 DoVs, which we take to be the identity of the robot and the identity of the task. However, we emphasize that this formalism can be extended to include variations like different objects, different environment dynamics, and so forth.
In our subsequent method description, we adhere to the regime specified above, where our universe has 2 DoVs: the robot and the task being performed. We use $R$ to denote the number of robots and $K$ to denote the number of tasks. The robots can have different system parameters, such as link length, configuration, geometry, and even state and action spaces of completely different dimensionality, with different numbers of joints and actuators. 
We assume that the $K$ tasks are achievable by all of the robots.

\subsection{Preliminaries}
\label{sec:prelim}
For each world $w$, let us define observations $o_{w}$ and controls $u_{w}$. The observations $o_{w}$ are the input that a robot would receive at test time, which could include images, encoder readings, motion capture markers, etc. In the case of complete information, the observations $o_{w}$ would be the full state $x_{w}$. The controls $u_{w}$ are the commands sent to the robot's motors, which could be joint torques, velocities, positions, etc. 
We assume access to a policy optimization algorithm that can perform policy search to learn a separate optimal policy $\pi_{w}$ for each world $w$.  A policy $\pi_{w}(u_{w}|o_{w})$ specifies a distribution over controls $u_{w}$ given an observation $o_{w}$. 
A policy search algorithm aims to search in policy space to find optimal parameters for $\pi_w$ which minimize the expected sum of costs $E_{\pi_w}(\sum_{t=0}^T c(o_{w}, t))$. Given an optimal policy $\pi^*_w$, we can draw actions $u_{w}$ given observations $o_{w}$ from the distribution $\pi^*_w(u_{w}|o_{w})$. 

For most worlds we consider, an observation $o_{w}$
can be split into a robot-specific ``intrinsic'' observation $o_{w, \robot}$
containing elements of the observation corresponding to the robot itself, and a task-specific ``extrinsic'' observation $ o_{w, \task}$ containing elements of the state corresponding to the task being performed. $o_{w, \robot}$ could include robot joint state and sensor readings, while $o_{w, \task}$ could include images, object locations, and the position of the robot's end-effector.
We assume that the state can be decomposed in the same way into $x_{w, \robot}$ and $x_{w,\task}$.
In order to decompose the policy by tasks and robots, we assume that the cost can be decomposed into a term that depends on the intrinsic state, and a term that depends on the extrinsic state: \mbox{$c(x_w,u_w) = c_\robot(x_{w, \robot}, u_w) + c_\task(x_{w,\task})$}, where the actions only affect the robot-dependent term, since they are inherently intrinsic.
This assumption is reasonable, as the cost tends to be in terms of object locations and torque penalties. This decomposition of states and observations is crucial to being able to learn modular policies, as explained in Section~\ref{sec:modularity}.

\subsection{Modularity}
\label{sec:modularity}
The problem we tackle is that of transferring information along values of each degree of variation while the remaining DoVs change. 
We intuit that for a particular degree of variation, all worlds with that DoV set to the same value can share some aspect of their policies. 
Going back to our 2-DoV universe as shown in Fig.~\ref{fig:possibleworlds}, 
with 3 DoF and 4 DoF arms, performing the tasks of opening a drawer and pushing a block, consider 2 of the possible worlds: $w_1$, which is a 3 DoF arm opening a drawer, and $w_3$, a 3 DoF arm pushing a block. Although these worlds require different strategies due to the different tasks being performed, the robots are the same and hence robot kinematics, and control dimensionality matches across both worlds. We hypothesize that, for a particular degree of variation, all worlds with that DoV set to the same value can share some aspect of their policies. This is achieved by making the policies modular, so that the policies for worlds $w_1$ and $w_3$ share a ``3 DoF arm'' part of the policy.

We let $\pi_{w_{rk}}(u|o)$ be the policy for the world $w$ with robot $r$ performing task $k$.
To make the notation clearer, let us say that $\pi_{w_{rk}}(u|o)$ is a distribution parametrized by a function $\phi_{w_{rk}}(o)$. For example, $\pi_{w_{rk}}(u|o)$ can be a Gaussian $\mathcal{N}(\phi_{w_{rk}}(o), \Sigma)$ with mean set to $\phi_{w_{rk}}(o)$, and $\phi_{w_{rk}}$ can be an arbitrary function on observations. 

For modular policies, we express $\phi_{w_{rk}}(o)$ as a composition of functions $f_r$ and $g_k$ that represent robot-specific and task-specific parts of the policy for robot $r$ and task $k$. Note that throughout our explanation, $f$ shall represent robot-specific modules and $g$ shall represent task-specific modules. These functions $f_r$ and $g_k$ act on the decomposed parts of the observation $o_{w, \robot}$ and $o_{w, \task}$ respectively. The compositionality of modules can be represented as  
\begin{equation}
    \phi_{w_{rk}}(o_w) = \phi_{w_{rk}}(o_{w, \task}, o_{w,\robot}) = f_r(g_k(o_{w, \task}), o_{w, \robot})
\end{equation}

We refer to $f_r$ as the robot-module for robot $r$ and the function $g_k$ as the task-module for task $k$. A separate set of parameters is used for each robot-module and task-module, such that worlds with the same robot instantiation $r$ would reuse the same robot module $f_r$, while worlds with the same task instantiation $k$ would use the same task module $g_k$. 

The reason the modules are composed in this particular order for the scenarios we consider is because we expect that the identity of the task would affect the task plan of the policy, while the robot configuration would affect the control output. An important point to note here is that the output of the task module $g_k$ is not fixed or supervised to have a specific semantic meaning. Instead it is a latent representation that is learned while training the modules.

If we consider a larger number of DoVs, such as robots, tasks, and objects on which those tasks are performed, we could arrange the modules in an arbitrary ordering, so long as the ordering forms a directed acyclic graph (DAG). In the general case, each module receives inputs from its children and the observation corresponding to its DoV, and the root module outputs the action.


This definition of modularity now allows us to reuse modules across worlds. Given an unseen new world $w_{\test}$, using robot $r_{\test}$ to perform task $k_{\test}$, modules $f_{r_{\test}}$ and $g_{k_{\test}}$, which have been learned from other worlds in the training set, can be composed to obtain a good policy. We do require that some world in the training set must have used robot $r_{\test}$ and some other world must have performed $k_{\test}$, but they need not have ever been trained together. For the unseen world, we have
\vspace{-0.2cm}
\begin{align*}
        \phi_{w_{\test}}(o_{\test}) &= \phi_{w_{\test}}(o_{w_{\test}, \task}, o_{w_{\test},\robot}) \\
        &= f_{r_{\test}}(g_{k_{\test}}(o_{w_{\test}, \task}), o_{w_{\test}, \robot})
\end{align*}


Note that we do not attempt to build a mapping relating different robots or tasks to each other, but instead use the experience of our desired robot on other task and the performance of our desired task with other robots to enable transfer of skills. As the number and variety of worlds which use a particular module increase, the module becomes increasingly invariant to changes in other DoV's, which is crucial for generalization. For example, as the number of robots being trained increases, each task module will need to work with various robot modules, which encourages them to become robot-agnostic. Similarly robot modules become task-agnostic as more tasks are performed with the same robot. 

\begin{figure*}[t!]
  \centering
  \includegraphics[ width=0.8\linewidth]{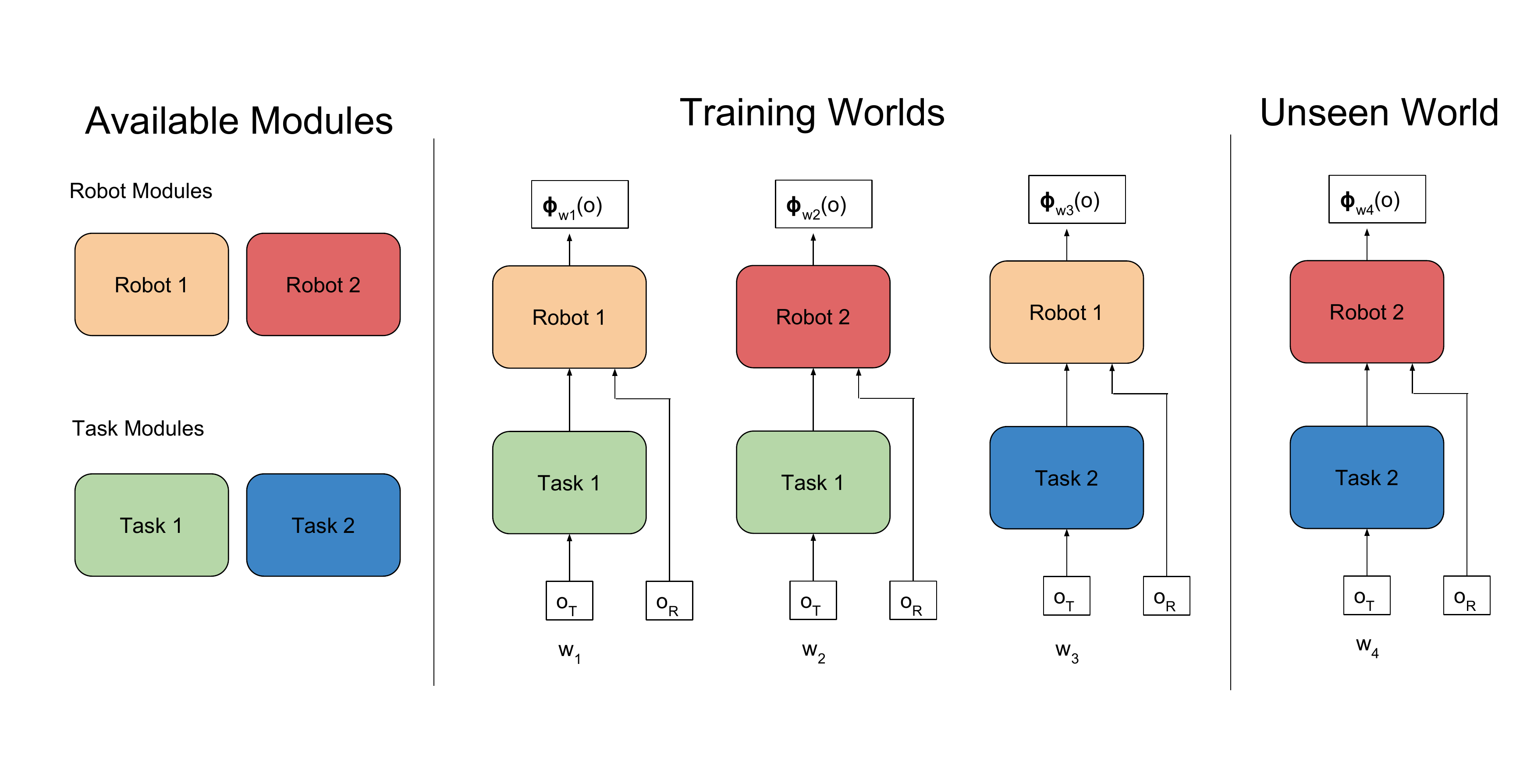}
  \vspace{-1cm}
  \caption{Modular policy composition for a universe with 2 tasks and 2 robots. There are 4 available modules - 2 task modules and 2 robot modules, and each module is a nerual network. For the training worlds, these modules are composed together to form the individual policy networks. Modules of the same color share their weights. Policy networks for the same task share task modules and those for the same robot share robot modules. The training worlds are composed and then trained end-to-end. On encountering an unseen world, the appropriate previously trained modules are composed to give a policy capable of good zero-shot performance}
  \label{fig:traingrid}
\end{figure*}


\subsection{Architecture and Training}
For this work, we choose to represent the modules $f$ and $g$ as neural networks due to their expressiveness and high capacity, as well as the ease with which neural networks can be composed. Specifically, for a world with robot $r$ and task $k$, we choose $\phi_{w_{rk}}(o)$ to be a neural network, and we choose the policy $\pi_{w_{rk}}(u|o)$ to be a Gaussian with mean set to $\phi_{w_{rk}}(o)$, and a learned but observation-independent covariance. Each policy mean is thus a composition of two neural network modules $f_r$, $g_k$, where the output of the task module is part of the input to the robot module.

Several training worlds are chosen with combinations of robots and tasks, such that every module has been trained for at least one world. This is illustrated in Fig.~\ref{fig:traingrid}.  

The combined grid of policy networks, with weights tied between modules, are trained using inputs from all the worlds. This is done synchronously, by first collecting samples from each of the worlds and them feeding them forward through their corresponding modules to output predicted controls for each world. However, asynchronous training methods could also be explored in future work.

Formally, training proceeds by minimizing the reinforcement learning loss function $\mathcal{L}$, which is the sum of individual loss functions $\mathcal{L}_w$ from each of the training worlds $w$. The details of the loss function and how it might be minimized, depends on the particular RL algorithm used to train the policies. In our experiments, we use guided policy search (GPS)~\cite{levinefinn16JMLR}, though other methods could be readily substituted. GPS proceeds by using local policy solvers to supervise the training of the final neural network policy, such that the loss for $\mathcal{L}_w$ is a Euclidean norm loss for regression onto the generated training actions. A more standard policy gradient might instead use the approximate linearization of the expected return as the loss~\cite{trpo}. Most policy learning methods, including GPS and policy gradient methods, require computing the gradient of $\log \pi(u|o)$ with respect to its parameters. In our method, as $\pi_{w_{rk}}$ is parametrized by a neural network $\phi_{w_{rk}}$ (with parameters $\theta_k$ for the task module and parameters $\theta_r$ for the robot module), we get the following gradients.
    \[\frac{\partial \pi_{w_{rk}}}{\partial \theta_r} = \frac{\partial \pi_{w_{rk}}}{\partial \phi_{w_{rk}}} \frac{\partial \phi_{w_{rk}}}{\theta_{r}}\]
    
    \[\frac{\partial \pi_{w_{rk}}}{\partial \theta_{k}} = \frac{\partial \pi_{w_{rk}}}{\partial \phi_{w_{rk}}} \frac{\partial \phi_{w_{rk}}}{\theta_{k}}\]
As $\phi_{w_{rk}} = f_{r}(g_{k}(o_{w, \task}), o_{w, \robot})$, we can rewrite the gradients as follows,
\[\frac{\partial \pi_{w_{rk}}}{\partial \theta_r} = \frac{\partial \pi{_{w_{rk}}}}{\partial f_r}\frac{\partial f_r}{\partial \theta_r}\]
\[\frac{\partial \pi_{w_{rk}}}{\partial \theta_k} = \frac{\partial \pi{_{w_{rk}}}}{\partial f_r}\frac{\partial f_r}{\partial g_k}\frac{\partial g_k}{\partial \theta_k}  \]

These gradients can be readily computed using the standard neural network backpropagation algorithm.


\subsection{Regularization}
In order to obtain zero-shot performance on unseen robot-task combinations, the modules must learn standardized interfaces. If, for example, a robot module overfits to the robot-task combinations seen during training, it might partition itself into different ``receptors'' for different tasks, instead of acquiring a task-invariant interface. With only a few robots and tasks (e.g. 3 robots and 3 tasks), we have found overfitting to be problematic. To mitigate this effect, we regularize our modules in two ways: by limiting the number of hidden units in the module interface, and by applying the dropout method, described below.

Limiting the number of hidden units in the outputs of the first module forces that module to pass on information compactly. A compact representation is less likely to be able to overfit by partitioning and specializing to each training world, since it would not be able to pass on enough information to the next module.

Dropout is a neural network regularization method that sets a random subset of the activations to 0 at each minibatch~\cite{dropout}. This prevents the network from depending too heavily on any particular hidden unit and instead builds redundancy into the weights. This limits the information flow between the task and robot modules, reducing their ability to overspecialize to the training conditions.

\section{Experiments}
To experimentally evaluate modular policy networks, we test our transfer approach on a number of simulated environments. For each experiment we use multiple robots and multiple tasks and demonstrate that using modular policy networks allows us to train on a subset of the possible worlds in a universe, and achieve faster or zero-shot learning for an unseen world. We evaluate our method against the baseline of training a separate policy network for each world instantiation. In order to evaluate our method on a challenging suite of simulated robotic tasks, we constructed several simulated environments using the MuJoCo physics simulator~\cite{tet-mjc-12}. We evaluate our method on tasks that involve discontinuous contacts, moving and grasping objects, and processing raw images from simulated cameras.For all experiments, further details and videos can be found at \url{https://sites.google.com/site/modularpolicynetworks/}

\subsection{Reinforcement Learning Algorithm}

The policy search algorithm we use to learn individual neural network policies is the guided policy search method described in ~\cite{levine2014learning}. 
This work splits policy search into trajectory optimization and supervised learning. To learn a number of local policies under unknown dynamics, the method uses a simple trajectory-centric reinforcement learning algorithm based on LQR. This algorithm generates simple local time-varying linear-Gaussian controllers from individual initial states of each system, with different controllers for different initial states. These controllers then provide supervision for training a global neural network policy using standard supervised learning, with a Euclidean loss regression objective. In our experiments, we use the BADMM-based variant of guided policy search which applies an additional penalty on the trajectory optimization for deviating from the neural network policy to stabilize the learning process~\cite{levinefinn16JMLR}. This choice of learning algorithm enables us to train deep neural networks with a modest number of samples. However, more standard methods, such as policy gradient \cite{williams92,trpo} and actor-critic \cite{ddpg, peters2008natural} methods, could also be used with modular policy networks.

\begin{figure}[t!]
  \centering
  \includegraphics[width=.75\linewidth]{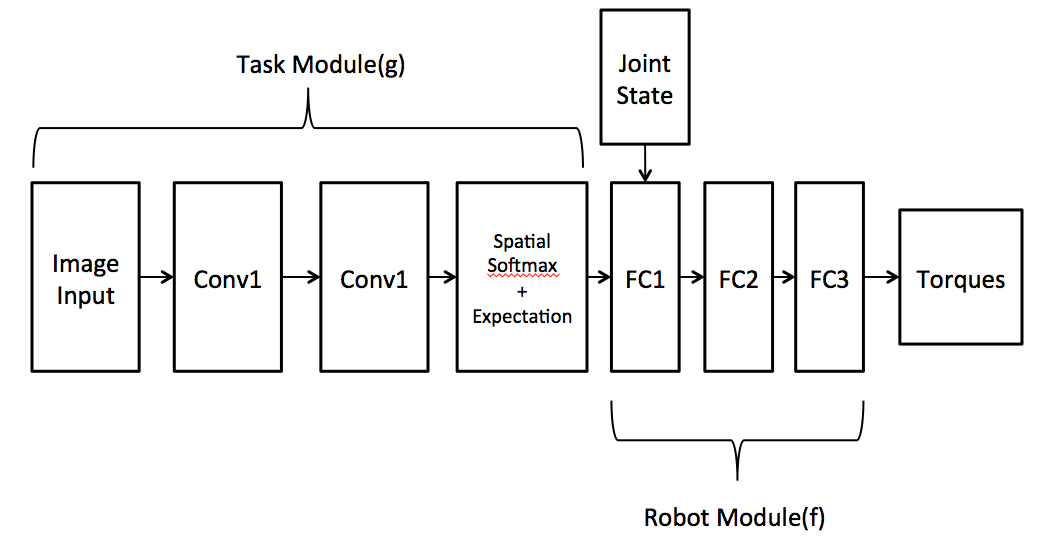}
  \caption{Basic visuomotor policy network for a single robot. The two convolutional layers and spatial softmax form the task module, while the last few fully connected layers form the robot module}
  \label{fig:basicnet}
\end{figure}

\subsection{Network Architecture}

For tasks that require performing simulated vision, we used a neural network architecture as shown in Fig.~\ref{fig:basicnet}. 
This architecture follows prior work~\cite{levinefinn16JMLR}. In non-vision tasks, the convolutional layers are replaced with fully connected layers. In both cases, the task module also takes as input the position of the robot's end-effector. Since the end-effector is present in all robots, we provide this to the task module so as to make it available to the policy in the earliest layers.




\label{sec:gridcolor}
\begin{figure}[t!]
  \centering
  \includegraphics[ width=.95\linewidth]{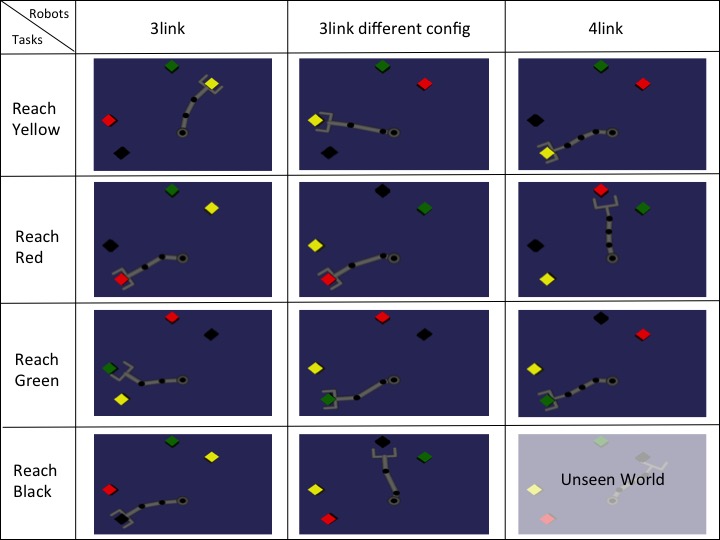}
  \caption{Grid of tasks vs robots for the reaching colored blocks task in simulation described in ~\ref{sec:colorblocks}. We train on all the worlds besides the 4link robot reaching to black, and test on this held out world.}
  \label{fig:gridcolor}
\end{figure}

\subsection{Reaching Colored Blocks in Simulation}
\label{sec:colorblocks}

In the first experiment, we evaluate a simple scenario that tests the ability of modular policy networks to properly disentangle task-specific and robot-specific factors in a visual perception task. In this task, the environment consists of four colored blocks (red, green, yellow, and black) positioned at random locations in the workspace, and each task requires the robot to reach for a particular colored block. The universe for this scenario includes three robots: a 3-link arm, a 3-link arm with links of a different lengths, and a 4-link arm. Each robot has its own robot module, and is controlled at the level of joint torques. The size of the image passed is 80x64x3, and the state space for each robots is its joint angles and its joint angle velocities. This results in 15366 inputs for the 3-link robots and 15368 for the 4-link robots.
An illustration of this task is shown in Figure~\ref{fig:gridcolor}.

Although this task is not kinematically difficult, it requires the task modules to pick up on the right cues, and the small number of tasks and robots makes overfitting a serious challenge. In order to evaluate zero-shot transfer in this setup, we train the modules on 11 out of the 12 possible world instantiations, with the 4 link robot reaching the black block being the unseen world. None of the other policies being learned can be trivially transferred to this world, as the 3 link robots have different dimensionality and the other task modules have never learned to reach towards other blocks. Successful transfer therefore requires perception and control to be decomposed cleanly across the task and robot modules. This is illustrated in Fig.~\ref{fig:gridcolor}.

We compare the performance of our method against two baselines: the first baseline involves executing a random policy, while the second involves running a policy learned for the 4 link robot but for a different colored block. These baselines are meant to test for trivial generalization from other tasks. The results, shown in Table~\ref{tab:color_blocks}, show that our method is able to perform the task well without any additional training, while the baselines have significant error. This illustrates that we are able to transfer visual recognition capabilities across robots, which is crucial for learning transferable visuomotor policies.

\begin{table}[]
    \centering
    \begin{tabular}{|c|c|c|c|} \hline
    Test Position & Random network & Wrong task module & \textbf{Ours} \\ \hline \hline
      1   & 1.16 & 2.34 & \textbf{0.12}   \\
      2   & 1.29 & 1.75 & \textbf{0.28}  \\
      3   & 1.35 & 1.65 & \textbf{0.21}  \\
      4   & 1.29  & 2.41& \textbf{0.08}  \\
            \hline
    \end{tabular}
    \caption{We evaluate the zero shot performance of the 4-link arm reach to the black block. The numbers shown in the table are average distances from the end-effector to the black block over the last 5 timesteps of a sample from the policy; a perfect policy would get 0. We see that composing the 4-link robot module with the reach to black-block task module generates very good performance (under the column Ours), while composing a different task module with the correct robot module, or running a random policy does quite poorly.
    \label{tab:color_blocks}}
\end{table}

\begin{figure}[t!]
  \centering
  \includegraphics[width=.95\linewidth]{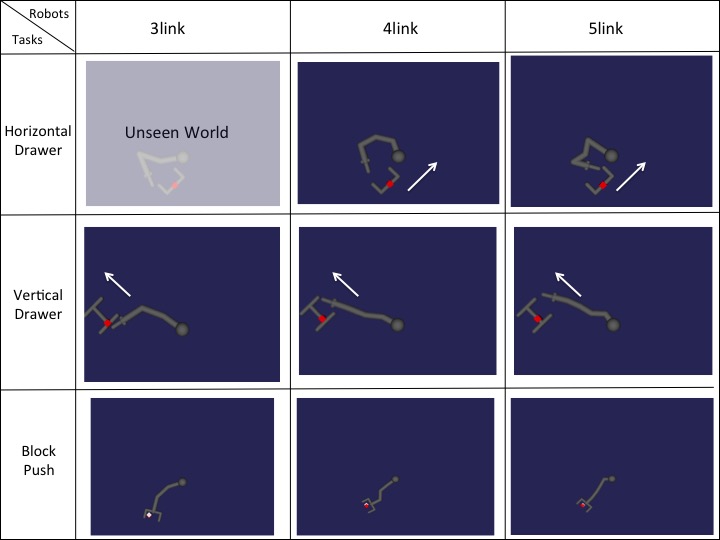}
  \caption{Grid of tasks vs robots for the object manipulation tasks described in ~\ref{sec:objectmanip}. The horizontal drawer tasks involve sliding the grey drawer horizontally to the target in the direction indicated by the arrow on the image. The vertical drawer tasks involve sliding the grey drawer vertically up in the direction indicated by the arrow on the image. The block push tasks involve pushing the white block to the red goal. All positions shown in this image are the final successful positions for each world.
  }
  \label{fig:grid2step}
\end{figure}


\subsection{Object Manipulation}
\label{sec:objectmanip}
In the next experiment, we evaluate our method on tasks that are more physically different to understand how well modular policy networks can transfer skills for manipulation tasks with complex contact dynamics. In this experiment, we use 3 robots and 3 tasks, as shown in Fig.~\ref{fig:grid2step}. The robots have 3, 4, or 5 links, with state spaces of different dimensionality, and we input target coordinates instead of images. The tasks are: pulling a drawer  horizontally, pushing a drawer vertically, and pushing a block to a target location. The arrows in the Fig.~\ref{fig:grid2step} indicate direction of motion for the tasks. Each of these tasks involve complex discontinuous contact dynamics at the point where the robot contacts the object. A grid of tasks and robots is presented in Fig~\ref{fig:grid2step}. In order to successfully transfer knowledge in this environment, kinematic properties of the tasks need to be transferred as well as dynamics of the robot.

\begin{figure}[t!]
  \centering
  \includegraphics[width=.75\linewidth]{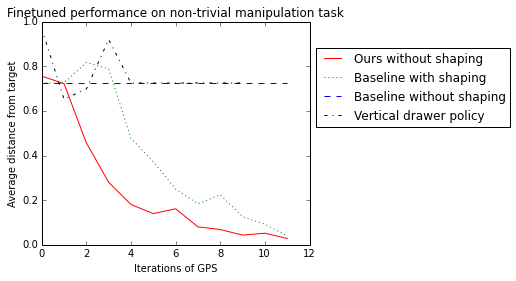}
  \caption{Results on the 3-link robot performing horizontal drawer pulling. The initialization from composing the 3-link robot module with the horizontal pull task module provides the fastest learning. Although the vertical drawer module was trained with the 3-link robot, the task is too different to directly transfer. Random initialization performs well with reward shaping, but without it is unable to learn the task.
 }
  \label{fig:manipresults}
\end{figure}

We train 8 out of the 9 possible worlds, with the 3 link robot pulling the horizontal drawer being held out. Although our method does not successfully perform zero-shot generalization directly simply by composing the modules for the held-out world, the transferred policy provides an excellent initialization for further learning. Figure~\ref{fig:manipresults} shows the learning curves with policies initialized using four paradigms: composing modules appropriately using our method, composing modules using the incorrect task-module (vertical drawer), and learning from scratch with and without shaping. In this task, the shaping term in the cost encourages the robot's gripper to reach for the drawer, while the standard cost without shaping simply considers the distance of the drawer from the target. Typically, tasks like this are extremely challenging to solve without shaping or a good initialization, since the robot must rely entirely on random exploration to learn to push the drawer before receiving any reward.

The results indicate that the transferred policy is able to learn the drawer task faster \emph{without} shaping than the task can learned from scratch \emph{with} shaping. When learning from scratch without shaping, the learning algorithm is unable to make progress at all. Therefore, if the shaping cost is not available, the policy obtained by transferring knowledge via modular policy networks is essential for successful learning. This indicates that, despite the wide variability between the tasks and robots and the small number of task-robot combinations, modular policy networks are able to transfer meaningful knowledge into the held-out world.

\begin{figure}[t!]
  \centering
  \includegraphics[width=.95\linewidth]{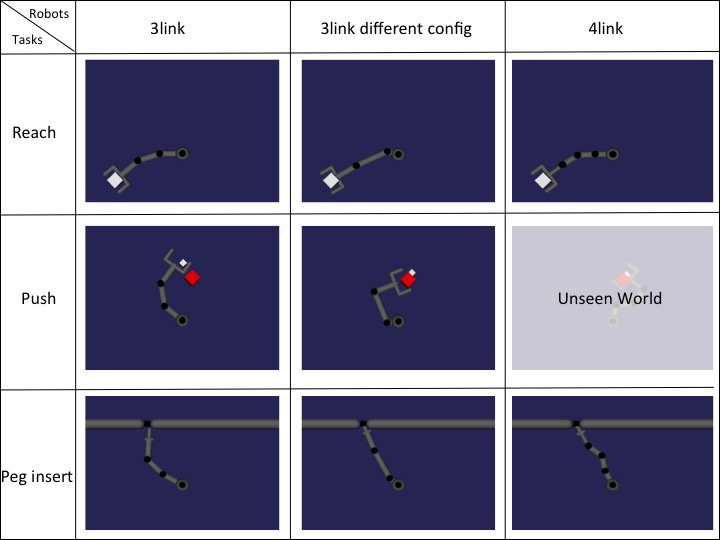}
  \caption{R-obots for visually distinct tasks mentioned in ~\ref{sec:visdiff}. The reach task involves reaching the white target. The push task involves pushing the white block to the red goal. The peg insertion task involves inserting the peg at the tip of the robot into the hole specified by the black square. These tasks involve manipulation and are also visually distinct in that they do not use the colors in the same way (the target and block colors are not consistent). We train on all combinations besides the 4link robot pushing the block.}
  \label{fig:gridvisuallydiff}
\end{figure}

\subsection{Visually Distinct Manipulation Tasks}
\label{sec:visdiff}

In the third experiment, we evaluated our method on a set of worlds that require both vision and physically intricate manipulation skills to succeed. An illustration of the tasks and robots in the experiment is presented in Figure~\ref{fig:gridvisuallydiff}. The robots again include the 3-link arms with different link lengths and a 4-link robot. The tasks require reaching to a given position, pushing a block to a given target, and inserting a peg into a hole. The goals for each task are visually distinct, and the tasks require a different pattern of physical interactions to handle the contact dynamics.

We choose 8 out of the 9 possible worlds to train, with the held out world being the 4 link robot pushing the block. This task is particularly difficult, since it involves discontinuous dynamics. Modular policy networks were able to succeed at zero-shot transfer for this task, significantly outperforming both a random baseline and policies from different robot-task combinations. This indicates that the modules were able to decompose out both the perception and the kinematic goal of the task, with the robot modules handling robot-specific feedback control to determine the joint torques needed to realize a given task.
\begin{table}[]
    \centering
    \begin{tabular}{|c|c|c|c|} \hline
    Test Position & Random network & Wrong task module & \textbf{Ours} \\ \hline \hline
      1   & 0.95 & 0.95  & \textbf{0.48}   \\
      2   & 1.79 & 1.14 & \textbf{0.19}  \\
      3   & 1.54 & 1.27 & \textbf{0.25}  \\
      4   & 0.94 & 1.32 & \textbf{0.23}  \\
            \hline
    \end{tabular}
    \caption{Zero-shot results on the 4-link performing the block-pushing task from section~\ref{sec:visdiff}. The values are the distance between the drawer and its target position averaged over tha last five time steps of each sample. Forming the policy by composing the 4-link module with the block pushing module performs best even those modules were not trained together. Choosing the reach module instead performs on par with a random network. We show that the task and robot modules are able to generalize to unseen robot-task combinations without additional training.
    \label{tab:visdiff}}
\end{table}

\begin{figure}[t!]
  \centering
  \includegraphics[ width=.75\linewidth]{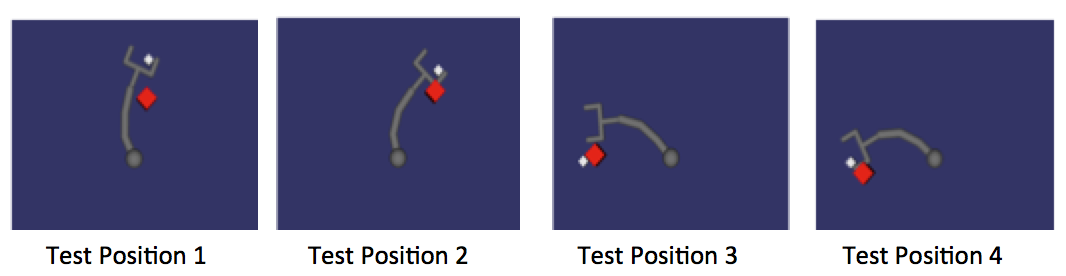}
  \caption{The final positions of the zero-shot performance of our method on the blockpushing task. Our method performs the task very well on zero shot and gets the white block to the red goal.}
  \label{fig:visdiffperformance}
\end{figure}

\section{Discussion and Future Work}

In this paper, we presented modular policy networks, a method for enabling multi-robot and multi-task transfer with reinforcement learning. Modular policy networks allow individual component modules for different degrees of variation, such as robots and tasks, to be trained together end-to-end using standard reinforcement learning algorithms. Once trained, the modules can be recombined to carry out new combinations of the degrees of variation, such as new robot-task combinations. The task-specific modules are robot-invariant, and the robot-specific modules are task-invariant. This invariance allows us to compose modules to perform tasks well for robot-task combinations that have never been seen before. In some cases, previously untrained combinations might generalize immediately to the new task, while in other cases, the composition of previously trained modules for a new previously unseen task can serve as a very good initialization for speeding up learning. 

One of the limitations of the current work is that, by utilizing standard reinforcement learning algorithms, our method requires different task-robot combinations to be trained simultaneously. In a practical application, this might require multiple robots to be learning simultaneously. A promising direction for future work is to combine our approach with more traditional, sequential methods for transfer learning, such that the same robot can learn multiple tasks in sequence, and still benefit from modular networks. This would enable combined lifelong and multirobot learning, where multiple robots might learn distinct robot-specific modules, trained sequentially, while contributing to shared task-specific modules, trained in parallel. By training on a larger variety of robots and tasks, the generalization capability of modular policy networks is likely to increase also. This could make it practically to automatically train large repertoires of different skills across populations of heterogeneous robotic platforms.



\bibliographystyle{IEEEtran}
\bibliography{references}

\begin{thebibliography}{10}
\providecommand{\url}[1]{#1}
\csname url@samestyle\endcsname
\providecommand{\newblock}{\relax}
\providecommand{\bibinfo}[2]{#2}
\providecommand{\BIBentrySTDinterwordspacing}{\spaceskip=0pt\relax}
\providecommand{\BIBentryALTinterwordstretchfactor}{4}
\providecommand{\BIBentryALTinterwordspacing}{\spaceskip=\fontdimen2\font plus
\BIBentryALTinterwordstretchfactor\fontdimen3\font minus
  \fontdimen4\font\relax}
\providecommand{\BIBforeignlanguage}[2]{{%
\expandafter\ifx\csname l@#1\endcsname\relax
\typeout{** WARNING: IEEEtran.bst: No hyphenation pattern has been}%
\typeout{** loaded for the language `#1'. Using the pattern for}%
\typeout{** the default language instead.}%
\else
\language=\csname l@#1\endcsname
\fi
#2}}
\providecommand{\BIBdecl}{\relax}
\BIBdecl

\bibitem{mnih}
V.~Mnih, K.~Kavukcuoglu, D.~Silver, A.~Graves, I.~Antonoglou, D.~Wierstra, and
  M.~A. Riedmiller, ``Playing atari with deep reinforcement learning,''
  \emph{CoRR}, vol. abs/1312.5602, 2013.

\bibitem{trpo}
J.~Schulman, S.~Levine, P.~Moritz, M.~I. Jordan, and P.~Abbeel, ``Trust region
  policy optimization,'' in \emph{International Conference on Machine Learning
  (ICML)}, 2015.

\bibitem{igordarwin}
I.~Mordatch, N.~Mishra, C.~Eppner, and P.~Abbeel, ``Combining model-based
  policy search with online model learning for control of physical humanoids,''
  in \emph{2016 {IEEE} International Conference on Robotics and Automation,
  {ICRA} 2016, Stockholm, Sweden, May 16-21, 2016}, 2016, pp. 242--248.

\bibitem{levinefinn16JMLR}
S.~Levine, C.~Finn, T.~Darrell, and P.~Abbeel, ``End-to-end training of deep
  visuomotor policies.'' \emph{Journal of Machine Learning Research}, vol.~17,
  pp. 1--40, 2016.

\bibitem{survey_deisenroth}
M.~P. Deisenroth, G.~Neumann, and J.~Peters, ``A survey on policy search for
  robotics,'' \emph{Found. Trends Robot}, vol.~2, no. 1\&\#8211;2, pp. 1--142,
  Aug. 2013.

\bibitem{survey_kober}
J.~Kober, J.~A. Bagnell, and J.~Peters, ``Reinforcement learning in robotics:
  {A} survey,'' \emph{I. J. Robotics Res.}, vol.~32, no.~11, pp. 1238--1274,
  2013.

\bibitem{reps_peters}
J.~Peters, K.~M{\"{u}}lling, and Y.~Altun, ``Relative entropy policy search,''
  in \emph{Proceedings of the Twenty-Fourth {AAAI} Conference on Artificial
  Intelligence, {AAAI} 2010, Atlanta, Georgia, USA, July 11-15, 2010}, 2010.

\bibitem{Schaal}
J.~Peters and S.~Schaal, ``Reinforcement learning of motor skills with policy
  gradients,'' \emph{Neural Networks}, 2008.

\bibitem{Taylor_stone_survey}
M.~E. Taylor and P.~Stone, ``Transfer learning for reinforcement learning
  domains: {A} survey,'' \emph{Journal of Machine Learning Research}, vol.~10,
  pp. 1633--1685, 2009.

\bibitem{Caruana95NIPS}
R.~Caruana, ``Learning many related tasks at the same time with
  backpropagation,'' in \emph{In Advances in Neural Information Processing
  Systems 7}.\hskip 1em plus 0.5em minus 0.4em\relax Morgan Kaufmann, 1995, pp.
  657--664.

\bibitem{Carauna97ML}
------, ``Multitask learning,'' \emph{Machine Learning}, vol.~28, no.~1, pp.
  41--75, 1997.

\bibitem{Crummon02JAIR}
C.~Drummond, ``Accelerating reinforcement learning by composing solutions of
  automatically identified subtasks,'' \emph{J. Artif. Intell. Res. {(JAIR)}},
  2002.

\bibitem{Konidaris06ICML}
G.~Konidaris and A.~Barto, ``Autonomous shaping: knowledge transfer in
  reinforcement learning,'' in \emph{International Conference on Machine
  Learning (ICML)}, 2006, pp. 489--496.

\bibitem{Ramon07ECML}
J.~Ramon, K.~Driessens, and T.~Croonenborghs, \emph{Transfer Learning in
  Reinforcement Learning Problems Through Partial Policy Recycling}, 2007.

\bibitem{Madden04AIR}
M.~G. Madden and T.~Howley, ``Transfer of experience between reinforcement
  learning environments with progressive difficulty,'' \emph{Artificial
  Intelligence Review}, vol.~21, no.~3, 2004.

\bibitem{GuestrinIJCAI03}
C.~Guestrin, D.~Koller, C.~Gearhart, and N.~Kanodia, ``Generalizing plans to
  new environments in relational mdps,'' in \emph{In International Joint
  Conference on Artificial Intelligence}, 2003.

\bibitem{pgella}
H.~B. Ammar, E.~Eaton, P.~Ruvolo, and M.~E. Taylor, ``Online multi-task
  learning for policy gradient methods,'' \emph{Journal of Machine Learning
  Research}, 2014.

\bibitem{konidaris07ijcai}
G.~Konidaris and A.~G. Barto, ``Building portable options: Skill transfer in
  reinforcement learning.'' in \emph{Proc. International Joint Conference on
  Artificial Intelligence}, 2007, pp. 895--900.

\bibitem{Mihalkova09IJCAI}
L.~Mihalkova and R.~J. Mooney, ``Transfer learning from minimal target data by
  mapping across relational domains,'' in \emph{Transfer Learning from Minimal
  Target Data by Mapping across Relational Domains}, 2009.

\bibitem{taylor07jmlr}
M.~Taylor, P.~Stone, and Y.~Liu, ``Transfer learning via inter-task mappings
  for temporal difference learning,'' \emph{Journal of Machine Learning
  Research}, vol.~8, no.~1, pp. 2125--2167, 2007.

\bibitem{Drummond02Jair}
\BIBentryALTinterwordspacing
C.~Drummond, ``Accelerating reinforcement learning by composing solutions of
  automatically identified subtasks,'' \emph{JAIR}, vol.~16, pp. 59--104, 2002.
  [Online]. Available: \url{http://jair.org/papers/paper904.html}
\BIBentrySTDinterwordspacing

\bibitem{Decaf}
J.~Donahue, Y.~Jia, O.~Vinyals, J.~Hoffman, N.~Zhang, E.~Tzeng, and T.~Darrell,
  ``Decaf: {A} deep convolutional activation feature for generic visual
  recognition,'' \emph{CoRR}, vol. abs/1310.1531, 2013.

\bibitem{domain_adaptation}
E.~Tzeng, J.~Hoffman, T.~Darrell, and K.~Saenko, ``Simultaneous deep transfer
  across domains and tasks,'' in \emph{International Conference in Computer
  Vision (ICCV)}, 2015.

\bibitem{andreas_modular_nets}
J.~Andreas, M.~Rohrbach, T.~Darrell, and D.~Klein, ``Deep compositional
  question answering with neural module networks,'' \emph{CoRR}, vol.
  abs/1511.02799, 2015.

\bibitem{ddpg}
T.~P. Lillicrap, J.~J. Hunt, A.~Pritzel, N.~Heess, T.~Erez, Y.~Tassa,
  D.~Silver, and D.~Wierstra, ``Continuous control with deep reinforcement
  learning,'' \emph{CoRR}, vol. abs/1509.02971, 2015.

\bibitem{minecraft}
J.~Oh, V.~Chockalingam, S.~P. Singh, and H.~Lee, ``Control of memory, active
  perception, and action in minecraft,'' \emph{CoRR}, vol. abs/1605.09128,
  2016.

\bibitem{Rusu16ArXiv}
A.~A. Rusu, N.~C. Rabinowitz, G.~Desjardins, H.~Soyer, J.~Kirkpatrick,
  K.~Kavukcuoglu, R.~Pascanu, and R.~Hadsell, ``Progressive neural networks,''
  \emph{CoRR}, vol. abs/1606.04671, 2016.

\bibitem{DBLP:Braylan15ArXiv}
A.~Braylan, M.~Hollenbeck, E.~Meyerson, and R.~Miikkulainen, ``Reuse of neural
  modules for general video game playing,'' \emph{CoRR}, vol. abs/1512.01537,
  2015.

\bibitem{dropout}
A.~K. I. S. R.~S. Nitish~Srivastava, Geoffrey~Hinton, ``Dropout: A simple way
  to prevent neural networks from overfitting,'' \emph{Journal of Machine
  Learning Research}, vol.~15, 2014.

\bibitem{tet-mjc-12}
E.~Todorov, T.~Erez, and Y.~Tassa, ``{MuJoCo}: A physics engine for model-based
  control,'' in \emph{International Conference on Intelligent Robots and
  Systems (IROS)}, 2012.

\bibitem{levine2014learning}
S.~Levine and P.~Abbeel, ``Learning neural network policies with guided policy
  search under unknown dynamics,'' in \emph{Advances in Neural Information
  Processing Systems}, 2014.

\bibitem{williams92}
\BIBentryALTinterwordspacing
R.~J. Williams, ``Simple statistical gradient-following algorithms for
  connectionist reinforcement learning,'' \emph{Machine Learning}, vol.~8, no.
  3-4, pp. 229--256, May 1992. [Online]. Available:
  \url{http://dx.doi.org/10.1007/BF00992696}
\BIBentrySTDinterwordspacing

\bibitem{peters2008natural}
J.~Peters and S.~Schaal, ``Natural actor-critic,'' \emph{Neurocomputing},
  vol.~71.

\end{thebibliography}

\end{document}